\documentclass[lettersize,journal]{IEEEtran}

\IEEEoverridecommandlockouts                              % This command is only needed if 
\usepackage{amsmath,amsfonts}
\usepackage{algorithmic}
\usepackage{algorithm}
\usepackage{hyperref}
\usepackage{array}
\usepackage[caption=false,font=normalsize,labelfont=sf,textfont=sf]{subfig}
\usepackage{textcomp}
\usepackage{stfloats}
\usepackage{url}
\usepackage{verbatim}
\usepackage{graphicx}
\usepackage{makecell}
\usepackage{cite}
\usepackage{xcolor}                             % Allows for setting text color
\usepackage{nicefrac}
\usepackage [autostyle, english = american]{csquotes}   % Fixes double quotes %
\MakeOuterQuote{"}  
\usepackage{siunitx}                            % Allows for properly formatted units
\usepackage{hyperref}
\title{\LARGE \bf
Soft Robotic Delivery of Coiled Anchors for Cardiac Interventions
}

\author{
Leonardo Zamora Ya\~nez$^{1}$, 
Jacob Rogatinsky$^{1}$, 
Dominic Recco$^{2}$, 
Sang-Yoep Lee$^{1}$, 
Grace Matthews$^{1}$,%
\\Andrew P. Sabelhaus$^{1}$, and Tommaso Ranzani$^{1}$%
\thanks{
*This work was supported by the National Institute of Biomedical Imaging and Bioengineering of the NIH, Awards R21EB028363 and R01EB035574.
The content is solely the responsibility of the authors and does not necessarily represent the official views of the NIH.
}%
\thanks{$^{1}$L. Zamora Ya\~nez, J. Rogatinsky, S.-Y. Lee, G. Matthews, A. Sabelhaus, and T. Ranzani are with the Department of Mechanical Engineering, Boston University, Boston, MA 02215, USA
    {\tt\small tranzani@bu.edu}
}%
\thanks{
$^{2}$D. Recco is with the Department of Cardiac Surgery, Boston Children's Hospital, Boston, MA 02115, USA
}%
}

\begin{document}
\maketitle
% \thispagestyle{plain}
% \pagestyle{plain}
% \thispagestyle{empty}
% \pagestyle{empty}

%%%%%%%%%%%%%%%%%%%%%%%%%%%%%%%%%%%%%%%%%%%%%%%%%%%%%%%%%%%%%%%%%%%%%%%%%%%%%%%%

\begin{abstract}
Trans-catheter cardiac intervention has become an increasingly available option for high-risk patients without the complications of open heart surgery. 
%
% Helical coils have been used to fix and implant various devices through various sized catheters.
%
However, current catheter-based platforms suffer from a lack of dexterity, force application, and compliance required to perform complex intracardiac procedures.
An exemplary task that would significantly ease minimally invasive intracardiac procedures is the implantation of anchor coils, which can be used to fix and implant various devices in the beating heart.
% through various sized catheters
% implant coiled anchors in a motile environment. 
%
% To address these limitations, w
We introduce a robotic platform capable of delivering anchor coils.
% similar to pacemaker leads and annuloplasty anchors.
%
We develop a kineto-statics model of the robotic platform and demonstrate low positional error. 
We leverage the passive compliance and high force output of the actuator in a multi-anchor delivery procedure against a motile \textit{in-vitro} simulator with millimeter level accuracy. 
\end{abstract}

%%%%%%%%%%%%%%%%%%%%%%%%%%%%%%%%%%%%%%%%%%%%%%%%%%%%%%%%%%%%%%%%%%%%%%%%%%%%%%%%

\begin{figure*}[t]
    \centering
    \includegraphics[scale=0.8]{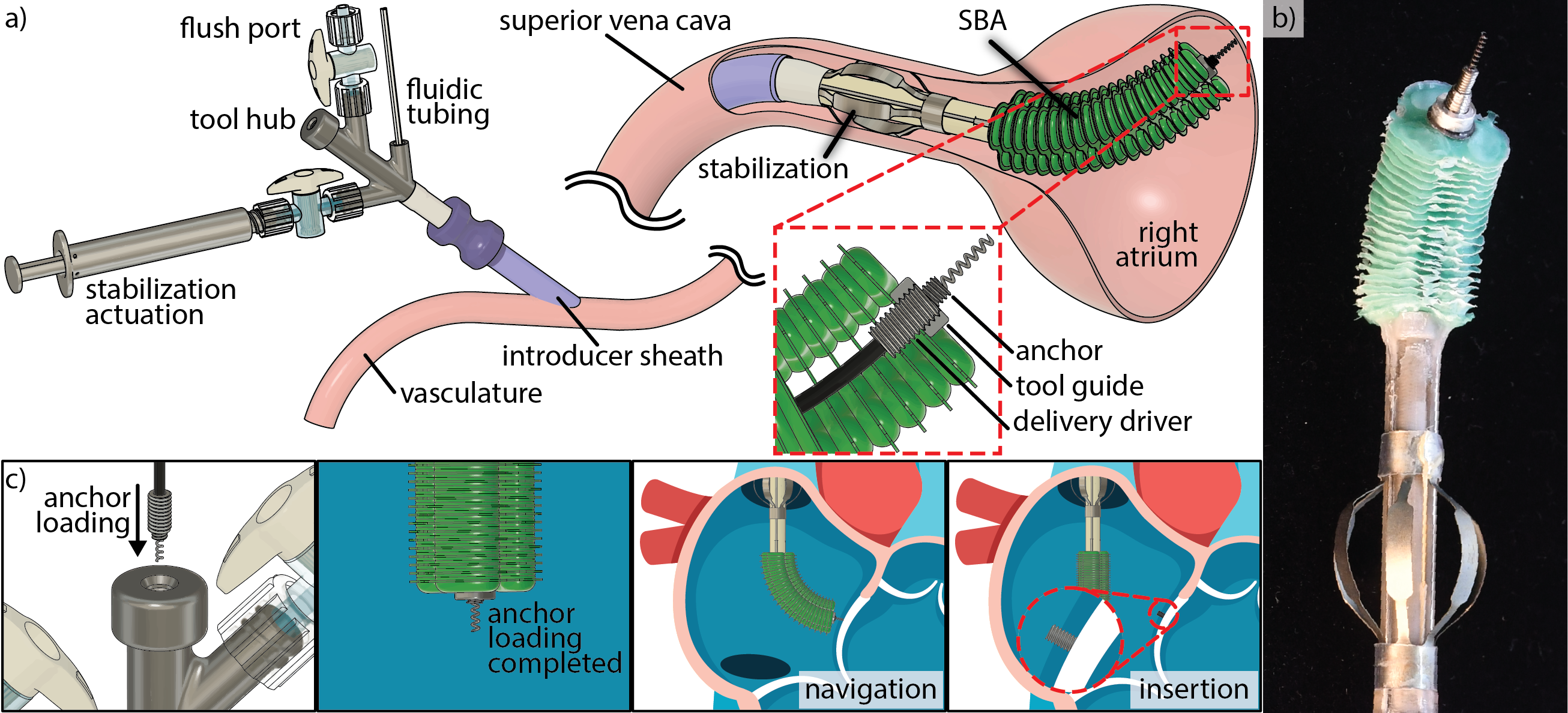}
    % \vspace{-2mm}
    \caption{
    \textbf{
    Robotic System Overview
    } 
    (a)~Robot overview from the proximal end (left) to the distal end (right). 
    (b)~Picture of the prototype with active stabilization and an anchor loaded at the tip.
    (c)~Loading and deployment of a coiled anchor to the tip of the robot for implantation.
    % inside the right atrium.
    }
    \label{fig:system_CAD}
    % \vspace{-5mm}
\end{figure*}

\section{Introduction}

Widespread access to minimally invasive, catheter-based, interventions for heart procedures would allow more patients to receive treatment for cardiac pathophysiology, without the risks associated with open-heart surgery (e.g., higher pain levels, longer hospital stays, and complications associated with cardiopulmonary bypass)~\cite{wang2017,mcdonagh2014,meduri2017,sadri2020}. 
Conventional catheters, while well suited for specific tasks such as guidance of implantable devices and therapeutic interventions, have limited dexterity and stability which makes it difficult to perform complex procedures~\cite{Ataollahi2016,Bechet2015,lee2017}.
%
% However, beating-heart trans-catheter procedures introduce challenges including constantly shifting workspace due to cardiac contractions and remote tool operation from a distant percutaneous access site~\cite{Ataollahi2016,Bechet2015,lee2017}.
%
% They are often preformed via a system of telescoping catheters with limited degrees of freedom (DOF). 
% \textcolor{red}{Coil implantation is an example of a task currently difficult with conventional catheters because hard to maintain contact with motile structures to ensure proper capturing of tissue, also difficult to position accurately.}
% \textcolor{red}{coil implantation is clinically significant: for example is needed for leadless peacemakers implantation which is used to threat arrythmias (stats you already wrote). Multiple coils can also be used to implant annuloplasty coils for treatment of valve regurgitation (stats and how many require open surgery).}
%
Anchor coil implantation is an example of a challenging task with conventional catheters due to the difficulty of maintaining stable contact with moving structures which is essential to capture tissue and position devices~\cite{russo2022}.
Coils can be used to implant active fixation pacemaker leads which are used to treat arrhythmias that affect over 10 million adults each year~\cite{noubiap2024, greenspon2012}.
Multiple coils can also be used to implant annuloplasty bands to treat valve regurgitation using minimally invasive techniques~\cite{Arnold2021, Chung2020}.
Valve regurgitation is a common valve pathology where leaflet coaptation is insufficient to prevent backflow during the cardiac cycle~\cite{hahn2023}.
Tricuspid valve regurgitation (TR) alone, affects over 70 million people worldwide, with more than 1.6 million cases in the US but interventions still lag behind with only 10,000 procedures preformed annually~\cite{stuge2006, messika2020}.
The majority of TR cases are treated via open surgery but these cases still retain a high mortality rate of 8.1\%~\cite{alqahtani2017} underscoring the need for new approaches.
Surgical robotic systems have emerged to address complexities in navigating tools to mobile anatomic targets while interacting with local structures with minimal visualization~\cite{peters2018, fagogenis2019}.
%
% Completion of cardiac interventions is difficult due to complexities in navigating tools to mobile anatomic targets while interacting with local structures with minimal visualization.
%
% Sequential probing with guidewires and catheters is often employed to localize target anatomy for device placement but this can lead to physician fatigue related to prolonged procedural times.
%
Robotic systems offer surgeons enhanced dexterity compared to traditional catheters while also enabling feedback to counter the indirect visualization experienced in minimally-invasive surgery~\cite{Wagner2006,Gang2011}.
Most robotic platforms for minimally invasive cardiac procedures focus on navigation and positioning for ablation and cardiac mapping~\cite{pappone2006, cruddas2021}.
Few systems have demonstrated complex procedures involving implantation of devices or tissue manipulation~\cite{machaidze2019,gosline2012}.
Among those that do, many rely on stiff, yet steerable, structures to transmit forces at the cost of introducing torsional instabilities restricting the range of possible procedures ~\cite{gosline2012,Vasilyev2013}.
Furthermore, their rigid design restricts the guidance of conventional instruments, often necessitating specialized end-effectors for procedures~\cite{yuen2013}
This rigidity can also hinder the robot's ability to navigate tortuous blood vessels while requiring complex computational systems to safely establish contact with the motile heart walls~\cite{yuen2013}.
To date, no robotic architecture addresses the challenges of coil implantation, highlighting the need for a trans-catheter system that can deliver coils with stable torque transmission and high accuracy in a dynamic heart environment.

We introduce a soft robotic platform for coil implantation~(Fig.~\ref{fig:system_CAD}) that addresses the following challenges: 
1) distal force transmission, 
2) navigation to moving targets,
3) continued safe contact with anatomic structures, and
4) accurate and reproducible anchor implantation.
%
% \textcolor{red}{I feel like you are missing some key contributions: you develop an anchor delivery system with integrated sensing; used the robot to position and deliver the anchors (exploit inherent complaince to maintain stable contact)}
%
We address the force transmission challenge by coupling the anchor to the robot’s tip to enable precise positioning and high force transmission.
% , enabling high force transmission required to puncture tissue
%
We introduce a coil delivery system with integrated sensing to detect when a self-release mechanism has fully embedded anchors of predefined length into \textit{in-vitro} and \textit{ex-vivo} environments. 
We then introduce a kineto-static model based on the soft robotic manipulator’s inflation to enable closed loop control.
We use this model to trace clinically relevant points along a path with millimeter accuracy.
We conclude by demonstrating how a combination of passive compliance by the robot and our closed-loop model can be used to position and and implant a series of nine anchors.
% with 100\% success rate.
% , thereby, minimizing operational time and complications.

\section{Design and Fabrication}
%
% The robotic platform can guide interventional catheters, such as those used in pacemaker implantation and annuloplasty procedures, inside the heart via its working channel.
%
% The robot features a threaded interface for high force transmission during implantation of coiled anchors.
% %
% We introduce a compact delivery handle with integrated sensing to deploy  anchors representative of those used in pacemaker leads and annuloplasty rings. 
% %
% This handheld device contains a self-release mechanism to ensure anchor deployment at the correct dept with one single motion.

\subsection{Surgical Robot}
The robot consists of a soft manipulator based on a Stacked Balloon Actuator (SBA)~\cite{Rogatinsky2022} and a stent-like stabilization mechanism which shifts the fulcrum point from the peripheral access port to the anatomical structures in close proximity to the heart (e.g., superior vena cava, Fig.~\ref{fig:system_CAD}(a)).
This robotic architecture was demonstrated to provide enhanced stability, dexterity, and force transmission for guiding interventional instruments in minimally invasive procedures inside the beating heart~\cite{Rogatinsky2023}.
% which is vital during coiled anchor delivery ~\cite{Rogatinsky2023}.
%
The SBA was designed with a central working channel able to accommodate tooling with a diameter of 3.2~mm (9.5~Fr) such as those used in this work for anchor delivery (Fig.~\ref{fig:system_CAD}(a)).
% , such as the anchor delivery driver (Fig.~\ref{fig:system_CAD}(a)). 
%
Three parallel stacks of balloons radially surround the working channel to enable 3-DoF control required for positioning the anchors.
% prior to delivery. 
%
Each stack is composed of twenty balloon layers with a maximum stroke of 40~mm at 100~kPa, sufficient to reach the His bundle for pacemaker delivery and the tricuspid valve for an annuloplasty.
The robot is actuated with saline to ensure compatibility with blood running though the vascular system.
% while providing high force transmission over long distances.

%
Given the significant displacement of valves during the cardiac cycle, upwards of 10~mm~\cite{singh2019}, coiled anchors must be securely held at the tip of the robot to prevent backwards slipping which would increase procedure error.
To address this challenge, we attached a threaded tool guide to the tip of the SBA for rigid coupling with the anchor delivery driver.
The delivery driver holds anchors and mates with the matching threads on the tool guide (Fig.~\ref{fig:system_CAD}(a)).
The threaded interface between the tool guide and delivery driver allows for accurate positioning of anchors, while preventing backward sliding of the delivery system, when engaging with moving anatomical structures.
The interface also enables high force transmission by leveraging SBA inflation to press the coiled end of the anchors against tissue for better capturing of the tissue during implantation.

The central channel terminates in a tool hub where instruments can be inserted through a hemostatic silicone valve (Fig.~\ref{fig:system_CAD}(a)). 
The fluidic lines, corresponding to each of the three SBA chambers, are connected to three custom syringe pumps.
Each syringe pump is driven by a stepper motor controlled by a motor driver~(TIC-T825, Pololu). 
User inputs are received through a keyboard and translated into actuator inputs through a microcontroller~(TEENSY40, PJRC) connected to a PC~(8NS97AV, HP) via the serial communication ports. 
%
% The microcontroller is connected to a PC~(8NS97AV, HP) via the serial communication ports.
% to implement path planning algorithms~(Algorithm~\ref{alg:pathTracing}).
%
The pressure in each chamber was measured using pressure gauges (P-7100-102G-M5, Nidec).
% to enable feedback for path planning. 

\subsection{Delivery System}
\begin{figure}[t]
    \centering
    \includegraphics[scale=1.0]{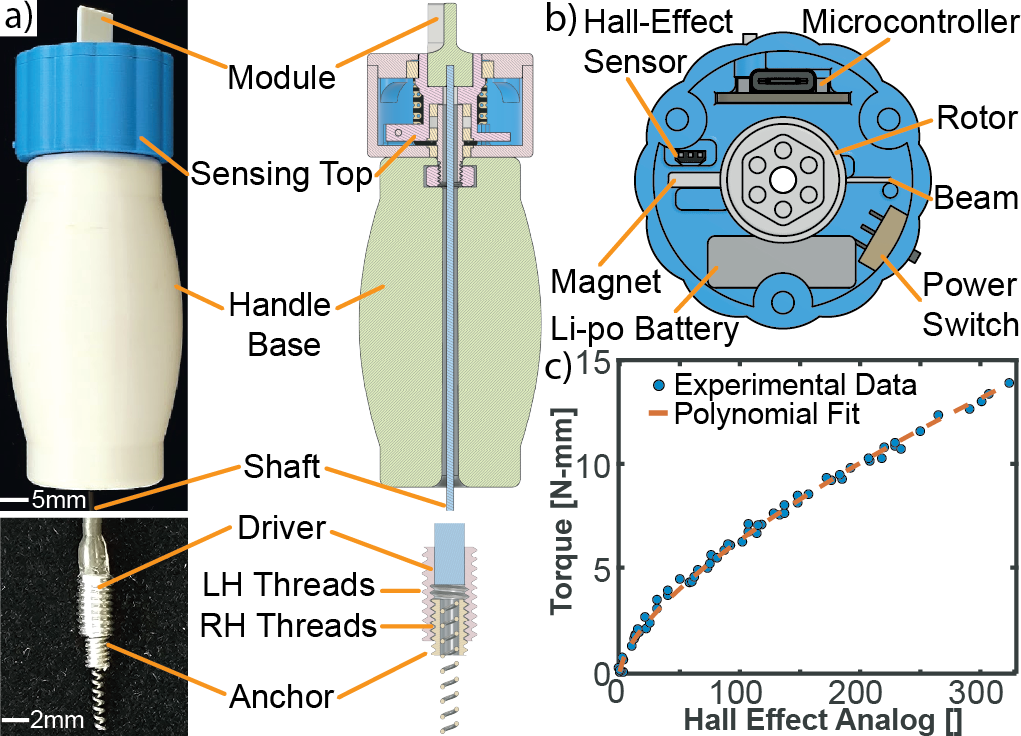}
    % \vspace{-5mm}
    \caption{
    \textbf{
    \label{fig:handle_delivery_sys}
    Anchor Delivery System
    } 
    (a)~Handle device overview including the replaceable module with a loaded anchor. 
    (b)~Electrical components embedded in the handle used for torque sensing. 
    (c)~Numerical model characterizing the magnetic torque sensing system. 
    }
    % \vspace{-5mm}
\end{figure}

To enable anchor delivery into cardiac tissue, a delivery system with onboard torque sensing was designed (Fig.~\ref{fig:handle_delivery_sys}).
The system consists of a palm sized handle, a modular driver, and onboard electronics for coil implantation torque sensing. 
The delivery system was designed to be operated with a single hand as interventional procedures often require bilateral movement coordination.
%
% , freeing the user’s second hand for separate operations. 
%
% This is important as interventional procedures often require bilateral movement coordination, thereby, mimicking other instruments commonly employed such as steerable sheaths and electrophysiology catheters. 
%
The handle measures 38~mm in diameter, 150~mm in height and weights 70~g. 
Although pacemaker electrodes only require one coil for implantation, annuloplasty bands require up to twenty coils for a complete procedure.
We address this challenge by designing an anchor delivery module from a braid reinforced catheter (CG412, Terumo Interventional Systems) with a magnetic interface on one end and a threaded driver on the other (Fig.~\ref{fig:handle_delivery_sys}(a)). 
The magnetic interface of the module is designed to allow for fast swapping once a new anchor is ready to be loaded.
The catheter was selected due to its high torque transmission while remaining flexible enough be steered by the robot.
The threaded driver features a cylindrical body with external M3$\times$0.5 left-hand threads and internal M2$\times$0.45 right-hand threads.
The internal threads enable coupling with the coiled anchors for deployment, while the external threads connect the driver-anchor assembly to the SBA via the tool guide.

The dual-thread driver enables self-release of anchors during deployment~(see supplementary video).
The self-release mechanism relies on a tissue dependent torque preload between the anchor and the driver. 
Once the anchor is loaded in the driver, the handle is operated by rotating the shaft counterclockwise which couples the anchor-driver assembly to the robot via the threaded tool guide.
The robot can then press the anchor coil against a surface, with further rotation driving the coil into the medium.
Once the coil is fully embedded, the anchor head contacts the medium which increases the torsional resistance of the driver.
Further rotation causes the driver-anchor threads to loosen, leading to their automatic decoupling.

Within the handle, the base was designed to rest in the palm of the user as the sensing top was rotated with their index or thumb (see supplementary video).
%
% To avoid using complex and delicate electrical interfaces, all electronic components were fit inside the sensing top of the handle. 
%
The electronic components fit inside the sensing top of the handle and include a hall effect sensor (DRV5055-Q1, Texas Instruments), a Bluetooth-equipped ESP32 microcontroller (Xiao ESP32C3, Seed Studio), and a battery with a power switch (Fig.~\ref{fig:handle_delivery_sys}(b)).
As torque is transmitted to the handle, an internal flexible beam bends, leading to an angular displacement of the rotor with a neodymium magnet. 
This rotation forces the magnet to approach the hall effect sensor which in turn increases the signal read by the microcontroller (Fig.~\ref{fig:handle_delivery_sys}(b)).
The sensor output was calibrated to match the torque experienced by a Force/Torque (F/T) sensor (9105-TW-NANO17-E, ATI) with a resolution of 0.03~N-mm, (Fig.~\ref{fig:handle_delivery_sys}(c)). 
A function mapping sensor signal to torque was derived and transmitted via Bluetooth to a computer for display.
The torque was estimated with an average error of 5\% and a nominal resolution of 0.07~N-mm.

\begin{figure}[t]
    \centering
    \includegraphics[scale=1.0]{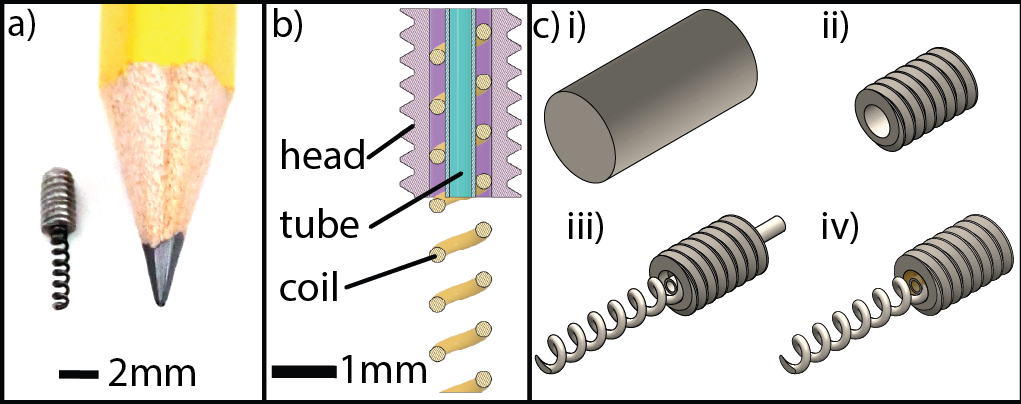}
    % \vspace{-5mm}
    \caption{
    \textbf{
    \label{fig:anchor_fab}
    Anchor Fabrication
    }
    (a)~Picture of fabricated anchor.
    (b)~Anchor fabrication steps.
    % starting from bar stock of 316 stainless steel. 
    (c)~Diagram of anchor components. 
    % Size comparison between an anchor and a \#2 wooden pencil.
    }
    % \vspace{-5mm}
\end{figure}

\subsection{Coiled Anchors}
Custom coiled anchors were designed and fabricated (Fig.~\ref{fig:anchor_fab}(a)).
% . as substitutes for pacemaker leads and annuloplasty anchors ~
%
% The size and geometry were determined by the dimensions reported in previous devices 0.7-1.0~mm coil diameter in pacemaker leads with a length of 6~mm in annuloplasty systems~\cite{zhao2011, mangieri2019}.
The size and geometry were based on devices used for pacemaker lead implantation and annuloplasty systems (i.e. 0.7-1.0~mm coil diameter, 6~mm in length)~\cite{zhao2011, mangieri2019}. 
% determined by the dimensions reported in previous devices 0.7-1.0~mm coil diameter in pacemaker leads with a length of 6~mm in annuloplasty systems~\cite{zhao2011, mangieri2019}.
%
Each anchor consists of three parts: a threaded head, a sharpened coil, and a pressure fit tube~(Fig.~\ref{fig:anchor_fab}(b)). 
The threaded head was fabricated on a lathe (HLV-H, HARDINGE) from 316L stainless steel bar stock (1335T43, McMaster)~(Fig.~\ref{fig:anchor_fab}(c)-i).
The stock material was turned down to 2~mm before threading the exterior surface using an M2$\times$.45 right hand die (a24011700ux0756, uxcell)~(Fig.~\ref{fig:anchor_fab}(c)-ii).
To accommodate the coil, a 1~mm through hole was drilled through the center of the anchor head.
The threaded end of the bar stock was then released to produce an anchor head with a diameter of 2~mm, length of 3~mm. 
%
% An off the shelf, 
A single helix coil (2924, Century Spring) was shortened to 8~mm before insertion into the anchor head 
A 304 stainless steel tube (5560K653, McMaster), with 0.4~mm diameter, was press fit inside the coil to secure the three part assembly ~(Fig.~\ref{fig:anchor_fab}(c)-iii). 
The pressed end of the coil was secured with medical adhesive (74795A74, McMaster) for added rigidity, and the free end was sharpened using a Dremel (4000-2/30, Dremel) equipped with a grinding wheel (Fig.~\ref{fig:anchor_fab}(c)-iv).

\section{Modeling and Control}
%
% While coil-based devices can be used in a variety of procedures, annuloplasty represents the most complex case and demands more sophisticated control strategies and is therefore the procedure used to evaluate the robot's performance.
%
To enable delivery of multiple coils along a trajectory, a requirement in minimally invasive annuloplasty procedures, we developed a path following algorithm for precise positioning of the robot.
%
% These demands include precise anchor positioning along a path following the a valve's annulus.
%
% Here we present an approach for enabling feedback control to follow a trajectory.
%
We first introduce a mechanics-based model to predict the height of the SBA's chambers under pressure.
We then employ a constant curvature framework to derive the kinematics of the robot and incorporate feedback from a 6-DoF sensor as part of the path following algorithm.

\subsection{Balloon Mechanics Model}
\begin{figure}[t]
    \centering
    \includegraphics[scale=1.0]{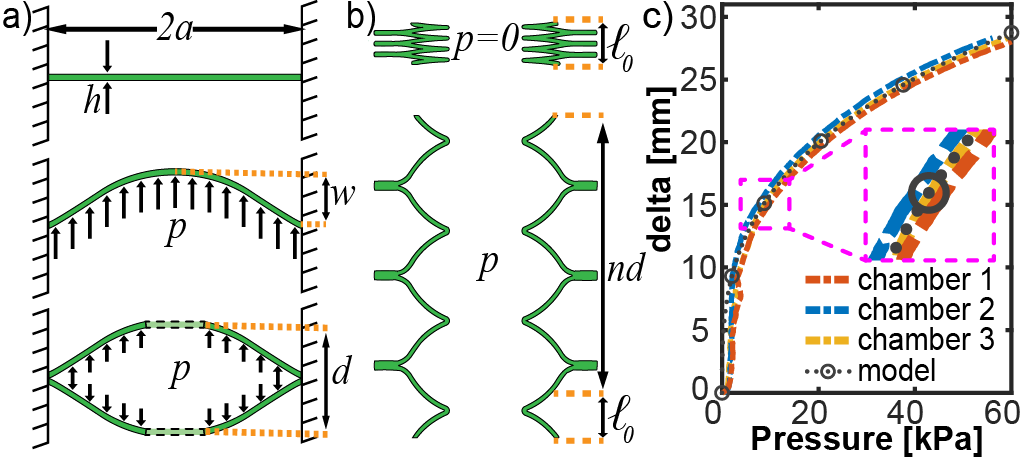}
    % \vspace{-4mm}
    \caption{
    \textbf{
    Kineto-Statics Model} 
    (a)~Timoshenko's plate model for large deflection compared to the model for a balloon. 
    (b)~Expanded model for a stack of balloons. 
    % (c)~Path followed by SBA for model characterization 
    (c)~Comparison of chamber inflation to analytical model.
    }
    \label{fig:mechanics_model}
    % \vspace{-5mm}
\end{figure}

The mechanics model predicts the maximum deflection of balloons composed of thin films under pressure.
% by incorporating both geometric and physical parameters
%
Unlike prior models, which focused on fully inflated balloons or relied on purely geometric estimations~\cite{paulsen1994,yang2020}, our approach is based on Timoshenko's theory for plates to account for large deflections in circular structures~\cite{timoshenko1959}. 
Plate theory describes the maximum displacement $w$ of a circumferentially clamped circular plate, as a function of a distributed load $p$, plate radius $a=4~mm$, thickness $h=0.038~mm$, and Young's Modulus at 50\% strain $E=13.4~Nmm^{-2}$ (Fig.~\ref{fig:mechanics_model}(a)).  

\begin{equation}
\label{eq:model_base}
w = 0.662a\sqrt[3]{ \frac{pa}{Eh} }
\end{equation}

This approach assumes $\frac{h}{w} << 1$, where $w$ is the deflection at the midpoint of the plate, and $h$ is the thickness of the plate.
We use $w=1.5~mm$ as a minimum deflection required for each balloon (total stroke of 30~mm for the 20 balloon SBA) resulting in $\frac{h}{w} = 0.025$.
% which meets the criteria for the model.

% to reach a total stroke of 30~mm.
% %
% We then arrive at the ratio  $\frac{h}{w_{0}} = 0.025$ which meets the criteria for the model $\frac{h}{w_{0}} << 1$.
% This approach assumes that deflection at the midpoint of the plate, $w_{0}$ is much larger the thickness of the plate, $h$.
% %
% We use $w_{0}=1.5~mm$ as a minimum deflection required for each balloon to reach a total stroke of 30~mm.
% %
% We then arrive at the ratio  $\frac{h}{w_{0}} = 0.025$ which meets the criteria for the model $\frac{h}{w_{0}} << 1$.
%
To model a balloon composed of two circular plates with open ends, we scaled Eq.(\ref{eq:model_base}) by a factor of two to account for the deflection of a second plate. 
A correction factor $c < 1$ was added to account for the open ends of the actuator balloons which will decrease the maximum deflection (Fig.~\ref{fig:mechanics_model}(b)). 
The model describes the maximum deflection $d$, of the balloon as function of pressure $p$, geometric parameters, and physical parameters of the balloon (Fig.~\ref{fig:mechanics_model}(a, b)).

\begin{equation}
\label{eq:model_balloon}
d(p) = 1.324c\sqrt[3]{ \frac{pa^{4}}{Eh} }
\end{equation}

The SBA consists of three parallel arrays of vertically stacked balloons. 
%
% Each stack shares a central channel which allows fluid to flow freely between neighbor- ing balloons. 
%
% Therefore, t
The vertical length, $\ell$, of $n$ stacked balloons due to a uniform load $p$ can be defined as a combination of deflated length $\ell_{0}$ and displacement $d$.

\begin{equation}
\label{eq:model_stack}
\ell(p) = nd_{0}(p) + \ell_{0}
\end{equation}

We verify Eq.(\ref{eq:model_stack}), by attaching a 6-DOF electromagnetic (EM) tracker (10006091 Aurora, NDI) to the tip of the SBA during inflation.
The three SBA chambers were then inflated with saline to a vertical tip displacement of 30~mm while recording the internal chamber pressure and measuring the vertical length with the EM probe. 
%
% The tracker was used to measure the vertical length of the SBA tip which was used as an approximation for the chamber heights. 
%
Fig.~\ref{fig:mechanics_model}(c) shows a plot of the pressure vs displacement of each chamber along with the analytical model.
% from Eq.(\ref{eq:model_stack}).
% can be seen in Fig.~\ref{fig:mechanics_model}(c)
%
The results show a maximum error of 1.34~mm for pressures over 5~kPa which corresponded to a 3.26\% error between the analytical model and experimental data. 
Below 5~kPa (12~mm displacement), the maximum error across all three chambers was found to be 7.82~mm. 
The larger error at lower pressures can be attributed to the stiction between the films that compose a balloon which lead to a lag the in inflation. 
%
% This phenomenon can be seen at the lower actuation pressures which lead to a lag in inflation of the SBA. 
%
In Fig. (\ref{fig:mechanics_model}(c)), the lag is manifested as a flattened section where an increase in pressure leads to near zero displacement until the threshold force to separate the layers is reached. 
%
% For the target procedures, much higher displacements, only achievable at higher pressures, are required, negating the downsides of the current model.

% \textcolor{red}{Please make sure that all components of Fig.~\ref{fig:mechanics_model} are cited in the text. I think part c is not. }

\subsection{Feedback control}
To enable navigation inside the heart, we built a  kinematics model based on the constant curvature framework~\cite{webster2010}. 
We identified the pressure input coordinates, $\mathrm{\underbar q}=[\ell_{1}(p_{1}), \ell_{2}(p_{2}), \ell_{3}(p_{3})]$.
Where $\ell_{i}$ and $p_{i}$ correspond to the length and pressure of the $i^{th}$ chamber.
Arc parameters arclenght $s$, curvature $\kappa$, and angle of the plane containing the arc $\phi$, were defined as functions of chamber lengths.
Using the relationships $\theta=\kappa s$ and $r=\frac{1}{\kappa}$, the tip position vector $\underbar r = [r\cos{\phi}(1-\cos{\theta}), r\sin{\theta}(1-\cos{\theta}), r\sin{\theta} ]$ was found. 
%
% \begin{equation}
% \label{eq:posVect}
% \mathrm{\underbar r} = [r\cos{\phi}(1-\cos{\theta}), r\sin{\theta}(1-\cos{\theta}), r\sin{\theta} ]
% \end{equation}
%

A numerical inverse kinematics solver was developed to trace clinically relevant paths along valve anatomy.
The solver was constructed by first obtaining the position Jacobian with respect to system pressures, $J = \frac{\partial \mathrm{\underbar r}}{\partial \mathrm{\underbar q}}$. 
The end tip position, $\underbar r$ was obtained from the 6-DoF sensor.
Next, we defined the error between a position along a path $\underbar g$, and the current tip position as $\underbar e = \underbar g_{n} - \underbar r$.
Where $g_{n}$ is the \textit{$n^{th}$} point along the path.
We then used the Jacobian pseudo-inverse $J^{\dag}$ and error vector, $e$, to obtain the required change in pressure inputs to minimize the error, Eq. (\ref{eq:gradient}). 
The error is further discretized using some small rate value $k < 1$ to minimize instability and maintain smooth actuation. 

\begin{equation}
\label{eq:gradient}
\Delta \underbar q = kJ^{\dag}\underbar e
\end{equation}

The following algorithm was constructed to numerically solve for the required pressure inputs to navigate through the path, $\underbar g$.
The error threshold, $\varepsilon$, was defined as the minimum distance between the robot tip and the goal point for the goal to be considered reached.

% MUST EDIT algorithm, include get apth points, get current path goal, etc
\begin{algorithm}
    \caption{Algorithm for path tracing}
    \begin{algorithmic}[1]
    \label{alg:pathTracing}
    \renewcommand{\algorithmicrequire}{\textbf{Input:}}
    \renewcommand{\algorithmicensure}{\textbf{Output:}}
    \REQUIRE $\underbar g$ - discretized path points
    % \ENSURE  $P_{i+1}$ - Set chamber pressures
    % \\ \textit{Initialization} :
    \WHILE { $n < length(g)$}
        \STATE Get current goal point, $\underbar g = \underbar v_{n}$
        \STATE Get chamber pressure vector, $\underbar P_{i}$
        \STATE Get tip position vector, $\underbar r_{i}$
        \STATE Calculate error $\underbar e = \underbar g_{n}-\underbar r_{i}$
        \IF { $||\underbar e|| < \varepsilon$  }
            \STATE Update goal point,  $\underbar g = \underbar g_{n+1}$
        \ELSE
            \STATE $\underbar P_{i+1}=\underbar P_{i} + kJ^{\dag}\underbar e$
            \STATE Set system pressures, $\underbar P_{i+1}$
        \ENDIF 
    \ENDWHILE
    % \\ \textit{LOOP Process}
    % \FOR {$i = l-2$ to $0$}
    % \STATE statements..
    % \IF {($i \ne 0$)}
    % \STATE statement..
    % \ENDIF
    % \ENDFOR
    % \RETURN $P_{i+1}$ 
    \end{algorithmic} 
\end{algorithm}

Based on the generated pressure inputs, the microcontroller used a PID controller to reach the desired chamber pressure using a syringe pump for actuation. 
%
% Then, the microcontroller returned the current chamber pressures to the PC which generated the next pressure inputs.  

\begin{figure}[hbtp]
    \centering
    \includegraphics[scale=1]{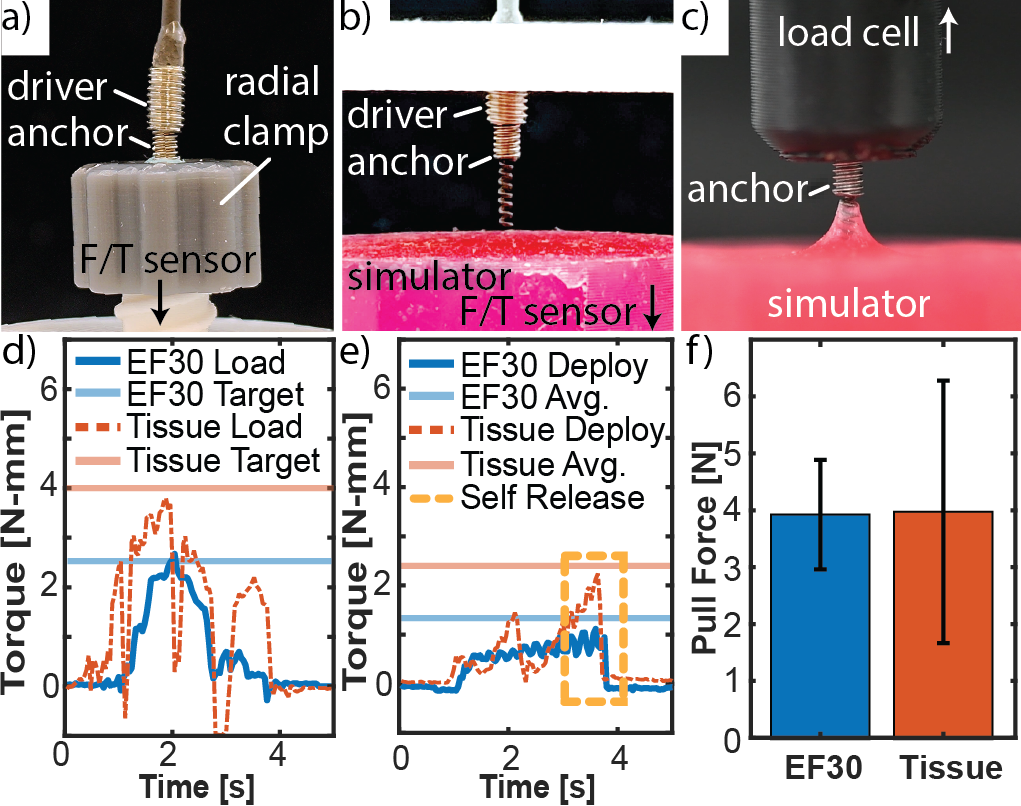}
    % \vspace{-5mm}
    \caption{
    \textbf{
    Deployment Characterization
    }
    (a)~Loading radially clamped anchor to threaded driver.
    (b)~Deploying anchor to \textit{in-vitro} simulator using driver and rigid alignment  plate.
    (c)~Pull force test for implanted anchor.
    % using a universal testing machine.
    (d)~Anchor loading sample for \textit{in-vitro} and \textit{ex-vivo} simulators.
    (e)~Anchor deployment sample for both simulators with highlighted drop in torque characteristic of the self-release mechanism.
    (f)~Average pull force for \textit{in-vitro} and \textit{ex-vivo} simulators.
    % demonstrating comparable tear properties between both mediums. 
    }
    \label{fig:anchor_char}
    % \vspace{-5mm}
\end{figure}

\section{Results}

\subsection{Deployment Characterization}
We characterize the anchor implantation process, and the stability of the anchors once implanted in three main steps.
% to ensure comparable performance to existing devices. 
%
% The evaluation consisted of three main steps.
%
First, the anchor is loaded into the driver with predetermined torque values depending on the target medium. 
Then, the anchor is implanted into a simulator to determine the torque required for full deployment.
Finally, the implanted anchor is pulled from the medium to measure the failure load.

%
% We designed two simulators, \textit{in-vitro} and \textit{ex-vivo}, to serve as mediums for implantation.
%
We validate the ability to implant coils \textit{in-vitro} and \textit{ex-vivo}.
The \textit{in-vitro} simulator consists of Ecoflex 00-30 (EF30) silicone (Smooth-On), selected for its comparable properties to porcine heart tissue~\cite{riedle2018determination, park2022development}.
The \textit{ex-vivo} simulator utilizes porcine tricuspid valve annulus tissue.
% clamped between two rigid plates.
%
Both designs allow for fresh samples to be inserted  between trials for consistent testing conditions. 
Each simulator is mounted to a F/T sensor to verify the loading and deployment torque exerted by the anchor delivery mechanism.

For each simulator, two sets of five anchors are used.
% in the characterization trials
%
Anchor loading began by radially clamping an anchor with a silicone ring to a rigid holder attached to the F/T sensor to measure the torque during loading~(Fig.~\ref{fig:anchor_char}(a)).
%
% The F/T sensor was used to measure the torque exerted on the anchor by the driver during loading. 
%
Each anchor was loaded to a pre-determined target torque value of 2.5~N-mm for EF30 and 4.2~N-mm for tissue (Fig.~\ref{fig:anchor_char}(d)).
%
% For the anchors used in these characterizations, the average loading torque was 2.55~N-mm for EF30 and 4.33~N-mm for tissue. 
%
Once the anchor reached the desired preload value, the driver-anchor assembly was released and prepared for deployment.

The deployment torque was defined as the maximum torque measured during the anchor implantation process. 
This value corresponded to the region where the self-release mechanism engaged, leading to slipping between the reverse threads and ultimately release of the anchor. 
To obtain this value, the driver-anchor assembly was vertically lowered onto a tissue simulator fastened to the F/T sensor until  0.5~N of force normal to the surface were measured~(Fig.~\ref{fig:anchor_char}(b)).
This value was established to ensure consistent puncture of the tissue by the coil required for full implantation. 
Torque values were recorded throughout deployment and continued until the self-release mechanism activated to implant the anchor (Fig.~\ref{fig:anchor_char}(e)). 
The average deployment torque was found to be 1.23~N-mm for EF30 and 2.55~N-mm for tissue.

\begin{figure}[t]
    \centering
    \includegraphics[scale=1]{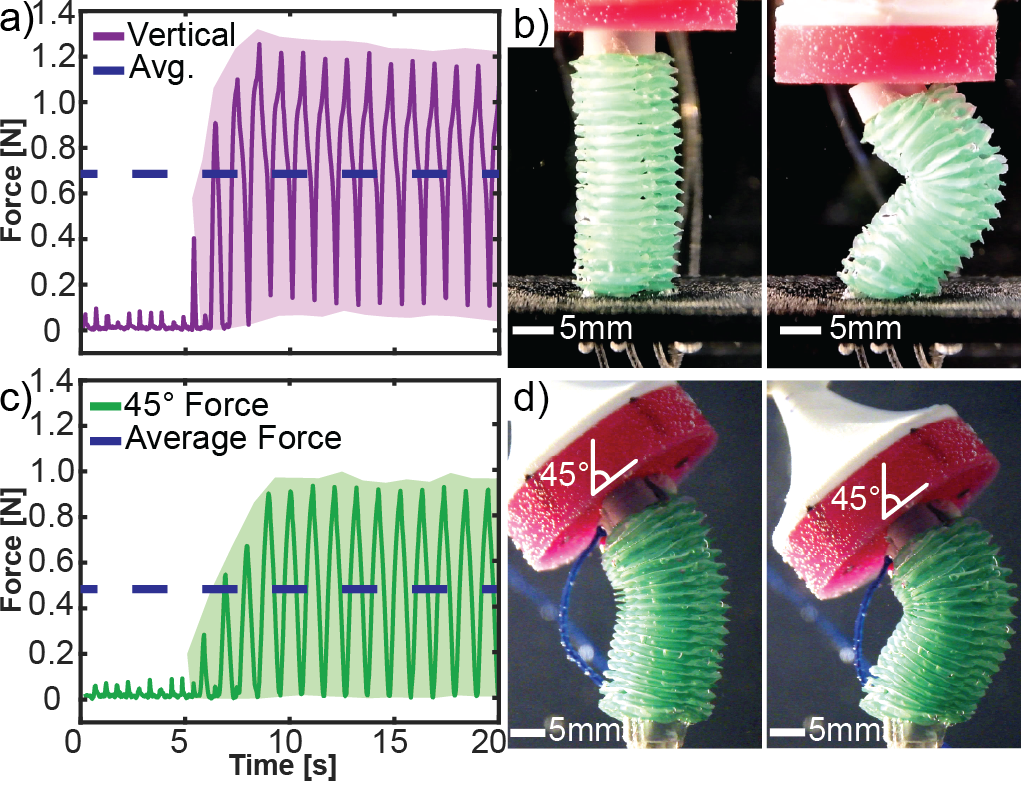}
    % \vspace{-6mm}
    \caption{
    \textbf{
    Robot Contact Force
    }
    (a)~Resultant force for vertical inflation of SBA against the moving simulator.
    (b)~Passive compliance against a motile \textit{ex-vivo} simulator.
    (c)~Resultant force for $45^{\circ}$ bending of SBA against the moving simulator.
    (d)~SBA inflation at $45^{\circ}$.
    }
    \label{fig:SBA_contact}
    % \vspace{-5mm}
\end{figure}

Following deployment, the maximum pull force of each anchor was measured.
% to ensure alignment with forces required during an annuloplasty.
%
Once an anchor was implanted, each sample was individually mounted to the base of a universal testing machine (5943, Instron) (Fig.~\ref{fig:anchor_char}(c)). 
The implanted anchor head was then coupled to the load cell using an M2 threaded adapter before raising it at a rate of $0.1~mms^{-1}$. %and pull force data was sampled at a rate of $50~Hz$.
The average maximum pull force for EF30 and tissue was $3.94~N$ and $3.99~N$ respectively which exceeds the required load of 0.52~N required for an annuloplasty~(Fig.~\ref{fig:anchor_char}(f))~\cite{adkins2015}.
%
% These results also supported our assumption that EF30 and porcine heart tissue have comparable mechanical properties. 
%
% We used this result to standardize the following experiments by using cast EF30 as a replica for a heart tissue simulator

\subsection{Robot Contact with Motile Structures}
%
% The right atrium is a blood-filed dynamic cavity which makes establishing contact with motile structures difficult.
%
To deliver anchors, the robot must exert adequate force and maintain constant contact while anchors are deployed into the tissue.
% at a small, localized area of interest.
%
We demonstrate how the robot can leverage its inherent softness to passively comply with a moving simulator while exerting the pre-established 0.5~N average force for anchor deployment. 
We choose to mimic the tricuspid valve annulus as the target anatomy due to its high positional variance, making it the most motile valve requiring coil implantation.

\begin{figure*}[t]
    \centering
    \includegraphics[scale=0.99]{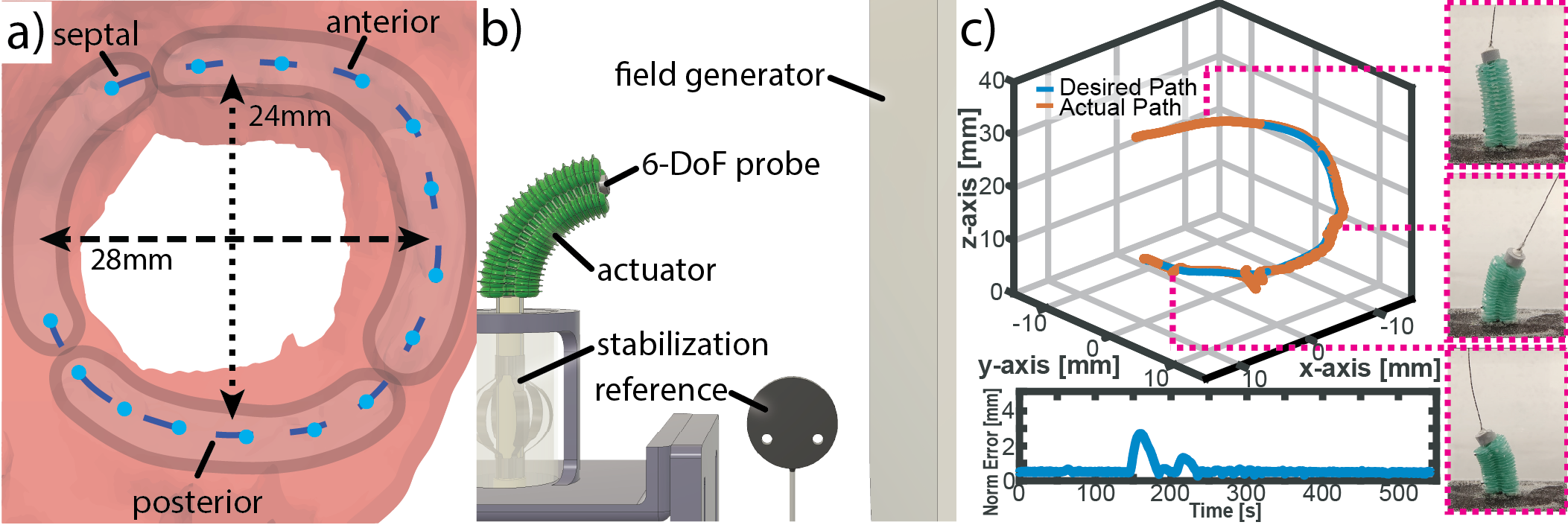}
    % \vspace{-3mm}
    \caption{
    \textbf{Path Tracing} 
    (a)~The path of the robot was obtained from a CT scan of the right atrium. The desired implantation locations were selected along the tricuspid valve annulus by a clinician.
    (b)~The device path tracing performance was evaluated underwater using an EM field generator and an EM probe to confirm tip position and error.
    (c)~The target locations were used to generate a smooth path in Euclidean space using spline functions. 
    The tip position of the actuator was measured using an EM tracker.
    } 
    \label{fig:pathTracing}
    % \vspace{-5mm}
\end{figure*}

% To replicate the challenges of a dynamic environment, a
An underwater motile simulator was constructed to evaluate the forces exerted on moving samples.  
The simulator body was fabricated by casting an EF30 sample on a circular base with an 80~mm extrusion. 
The extrusion served as an offset to allow for movement underwater without damaging sensing components. 
The simulator was fastened to the tri-axial F/T sensor and mounted to the end effector of a UR5e Universal Robot Arm.
% to receive tri-axial force data. 
%
% The simulator-sensor assembly was then mounted to the end effector of a UR5e Universal Robot Arm.
%
We programmed the arm to cycle 8~mm normal to the circular EF30 surface which corresponds to the upper extent of movement for the tricuspid valve annulus~\cite{owais2014}. 
The movement period was set to $60~Hz$ which corresponds to a normal heart rate of an adult undergoing cardiac surgery~\cite{luo2023}. 
The simulator was then submerged underwater.
% until the entire EF30 surface was covered.
%
% The simulator was then submerged underwater until the entire EF30 surface was covered. 
%
The robotic system was then positioned underwater, 40~mm from the simulator.
% 
% Once the robot was in position, t
The stabilization system was then engaged to secure the base of the actuator against a 35A silicone tube~(5236K527, McMaster).
%

%
% To asses the robot's ability to maintain contact with the beating heart during coil implantation, 
A right atrium CT scan was used to determine the bending angles required to reach the entirety of the tricuspid valve annulus, where coils need to be implanted for an annuloplasty.
An interventional cardiologist marked target points along the tricuspid valve annulus which were then translated into geometric angles for the robot. 
From the target points, we determined that the robot should contact the simulated annulus at angles between 0$^\circ$ (Fig.~\ref{fig:SBA_contact}(b)) and 45$^\circ$ (Fig.~\ref{fig:SBA_contact}(d)) from the axis normal to the actuator base.
%
% To account for this range, the two extremes were selected as the angles in the following force contact test.
%
% The volumes required to reach these angles were pre-determined, 0.48~mL for all three chambers in the vertical inflation case and  0.52~mL, 0.46~mL, 0.40~mL for the 45$^\circ$ case.  

For each angle, five trials were conducted to evaluate the contact force on the motile simulator.
Force data was collected for 20~s, starting with the deflated SBA and ending after ten cycles of contact with the simulator.
% 
% Force data was collected at using the F/T sensor a rate of $25~Hz$.
% to capture force oscillations. 
%
During the vertical loading case, the SBA exerted an average of  $0.63~N$  to the silicone simulator~(Fig.~\ref{fig:SBA_contact}(a)).
For the 45$^\circ$ case, the SBA exerted an average of $0.44~N$ (Fig.~\ref{fig:SBA_contact}(b)). 
The maximum forces for the 45$^\circ$ and vertical trial were found to be $1.0~N$ and $1.32~N$ respectively.
These forces fall below 5.5~N which is reported as the cutoff for damaging tissue~\cite{yuen2013} but are within range of the required force to implant coiled anchors.
The variance of the forces during contact can be attributed to the large displacement of the \textit{in-vitro} simulator.

\subsection{Path Tracing}
Semi-autonomous navigation
% has shown significant benefits when navigating robotic devices inside the beating heart and 
can be beneficial to implant multiple anchors along a predefined path as required in an annuloplasty~\cite{fagogenis2019}.

We demonstrate the efficacy of our path-following algorithm to trace the outline of the tricuspid valve annulus. 
The labeled CT scan was used to establish 15 anchor implantation locations along the annulus, spaced 5~mm apart (Fig.~\ref{fig:pathTracing}(a)).
The path was generated by calculating a series of 3D- splines that would sequentially connect all 15 points.
Once all the curves were generated, the entire path was discretized into 500 equally spaced points.
%
% The final path resembled an ellipse with a major diameter of 28~mm and minor diameter of 24~mm.
%
The robot was submerged underwater and the stabilization mechanism was engaged into silicone tubing to brace the base of the actuator  (Fig.~\ref{fig:pathTracing}(b)).
A 6-DOF EM position sensor (610164, Northern Digital Incorporated) was integrated at the SBA tip to receive position data. 
% To enable feedback, an EM tracking system was used (10006091, Northern Digital Incorporated). 
%
% Given its application in cardiovascular surgery, 
The EM field generator was positioned at a clinically relevant distance of 20~cm from the robot, which exceeds the typical distance, 10~cm, from the right atrium to the surface of the chest wall~\cite{seth2002}.
%

%

% Once the device was in position, the system was connected to a desktop PC to make use of the path following algorithm.
%
The pre-generated path was then loaded and stepper motor commands were generated to move the actuator tip towards the first target point (Fig.~\ref{fig:pathTracing}(c)).
After the SBA reached the first point, tip position data was recorded to estimate the error between the actual and desired position.
% across the entire path.
%
Along the path, the median error was 0.51~mm with a minimum and maximum error of 0.27~mm and 2.70~mm respectively~(Fig.~\ref{fig:pathTracing}(c)).

\subsection{Coiled Anchor Robotic Implantation}
% we now demonstrate the robot-assisted delivery of coiled anchors in a motile \textit{in-vitro} heart environment.
%
To verify the performance of the robot and delivery system in anchor implantation, we performed robot-assisted anchor deployment.
% using the same motile 
% simulator detailed in Sect.~\textcolor{red}{XXX}.
% used in 
% the force tests was used for 
%
A series of anchors were deployed into the \textit{in-vitro} motile simulator described in Sect.~IV.B.
Each anchor was individually loaded onto the driver using a torque cutoff of 2.5~N-mm. 
The delivery module was then attached to the handle and secured via the magnetic connecting interface  (Fig.~\ref{fig:handle_delivery_sys}). 
% %
% The anchor end of the handle was inserted through the working channel of the robot until it reached the distal tip. 
% %
% Once at the tip, the driver was coupled to the tool guide via the threaded interface.
% 

\begin{figure*}[t]
    \centering
    \includegraphics[scale=0.35]{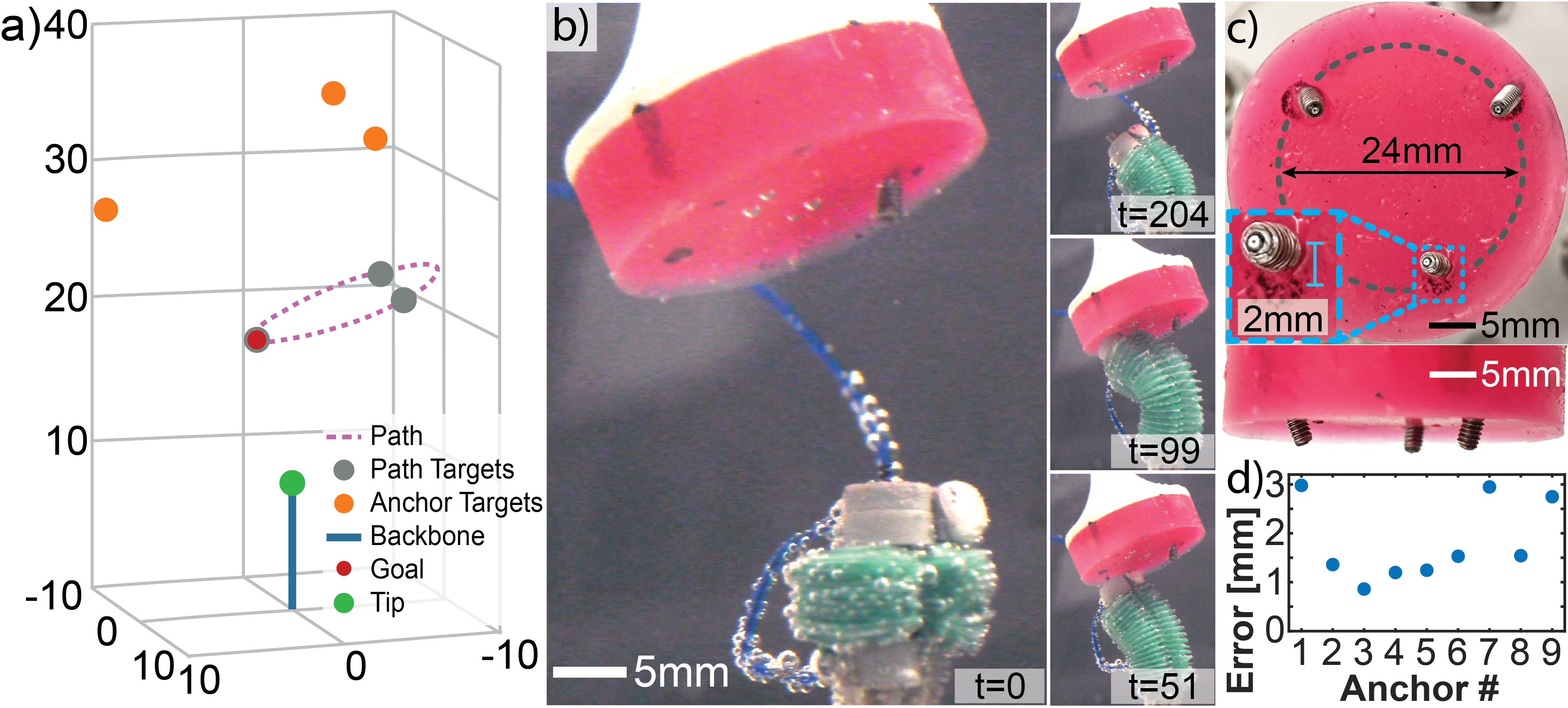}
    % \vspace{-3mm}
    \caption{
    \textbf{Anchor Implantation}
    (a)~User interface displaying the robot backbone, anchor targets on a motile simulator and targets along a circular path. 
    (b)~Sequential images of the robot delivering anchors into a motile simulator. At t=0, the robot begins with a loaded anchor. Then the path-planning algorithm is engaged to orient the robot towards the target location. The user then make the necessary adjustments to contact the motile surface before implantation. Finally, the user rotates the delivery system to implant the anchor.
    (c)~Closeup view of deployed  anchors in the \textit{in-vitro} simulator with a side view demonstrating full implantation.
    (d)~Error plot of all nine anchors delivered.
    } 
    \label{fig:anchorImplantation}
    % \vspace{-5mm}
\end{figure*}

%
We generated a circular path with radius 24~mm and with three anchor implantation locations equally spaced. 
The three anchor targets were marked onto the \textit{in-vitro} simulator.
By attaching a 6-DoF tracker onto the simulator, we obtained the position of the moving anchor targets.
%
% With these positions, 
A user interface was developed that displayed the anchor targets, the SBA tip position, and three new targets along a path projected from the moving simulator~(Fig.~\ref{fig:anchorImplantation}(a)).
These projected path targets served as goal targets to be reached by the close loop path planner prior to making contact with the moving \textit{in-vitro} simulator.
This approach positioned the SBA in the correct orientation using Algorithm 1 which enabled the user to contact the desired target with minimal adjustments.

The procedure for anchor implantation was as follows. 
First the user loaded an anchor into the SBA tip via the central lumen and rotated the driver to ensure coupling to the tool guide via the threaded interface, (Fig.~\ref{fig:anchorImplantation}(b)).
Then the path planning algorithm was leveraged to move the robot to one of the projected points along the path which would orient the robot towards the target location. 
Once the close loop controller reached the desired location, the user regained manual control to establish contact with the motile simulator. 
The user visually confirmed contact with the user interface and made any final adjustments.
% via visual confirmation. 
%
Once the user contacted the desired anchor target, the driver was rotated to embed the anchor into the simulator.
Nine anchors where implanted and their respective errors were measured using image analysis tools (Fig.~\ref{fig:anchorImplantation}(d)). 
The final mean error for anchor implantation was 1.82~mm with a minimum and maximum error of 0.86~mm and 2.92~mm, respectively.

\section{Conclusions}
Anchor coil implantation is a critical step to enable minimally invasive approaches for beating heart procedures and it is highly relevant to common procedures such as pacemaker lead delivery and treatment of valve regurgitation.
We present a soft robotic platform for minimally invasive implantation of coiled anchors.
Our platform uses a stabilization mechanism to locally brace the base of the robot while utilizing the inherent softness of the SBA to passively comply against a moving surface.
We developed a coil anchor delivery system that is compatible with a typical surgical workflow.
The delivery system uses a system of threads to 1) hold a coiled anchor with a threaded driver and 2) rigidly couple the driver to the tip of the robot for high force transmission, precise coil positioning, and to prevent tool slipping.
The system also features a self-release mechanism to fully implant coils into \textit{in-vitro} and \textit{ex-vivo} simulators without adding procedure complexity.
The threaded coupling with the SBA allowed for effective force transmission, sufficient to penetrate the tissue and implant the anchor.
%
% During deployment, a sharp drop in torque was measured and identified as a marker of the self-release mechanism.
%
After deployment, the average pull force in the \textit{in-vitro} and \textit{ex-vivo} simulators was measured, 3.94~N and 3.99~N respectively, which outperforms the forces required for cinching a valve during an annuloplasty.

To address the need for multiple coil implantation, we introduced a path following algorithm.
We derived a mechanics-based model for the actuator chamber lengths to analytically predict the displacement of a pressurized stack of balloons within 1.34~mm for actuation pressures above 5.0~kPa.
We demonstrate how chamber inflation can be used to derive a kineto-statics model and we show close loop control exploiting a 6-DOF tracker.
% to enable  of the robot. 
%
The performance of the controller was evaluated by tracking a trajectory derived from a tricuspid valve CT achieving an median norm error under
0.51~mm in \textit{in-vitro} experiments. 
We demonstrated how this robotic architecture can be used to enable path planning for sequential anchor placements, a key characteristic of an annuloplasty procedures. 
These results show promise in reducing the burden of navigation in the beating heart, a procedure that, while challenging, is time consuming and secondary to precise implantation.

We concluded by demonstrating how a combination of the delivery mechanism, close-loop controller, and robot compliance, can be used to accurately deliver coiled anchors into a motile target. 
Once the anchor was loaded and coupled to the robot, the path planning algorithm was used to position the system towards the goal site. 
Final adjustments were made by the user through the user interface and visual confirmation before implanting anchors into the simulator. 
This procedure enabled precise anchor placement with a position norm error of 1.82~mm and 100\% implantation success rate of nine anchors at a predefined depth of 5~mm.
These results demonstrate the millimeter level precision required for implantation into common coil targets, such as the His bundle and tricuspid annulus.

Future efforts will focus
% on improving system control by integrating force sensing at the tip to allow for active compliance.
%
integration with medical imaging such as ultrasound to move towards \textit{in-vivo} validations.

\end{document}